\documentclass{article}
\usepackage{iclr2027_conference,times}
\usepackage[T1]{fontenc}
\usepackage[utf8]{inputenc}
\usepackage{amsmath,amssymb,graphicx,booktabs,tabularx,microtype}
\usepackage{caption,placeins,float,needspace}
\usepackage{hyperref,url}
\hypersetup{hidelinks,pdftitle={From Reward Signal to Visual Utility: A Controlled Audit of Medical VLM Post-Training},pdfauthor={Wang Jingxin}}
\iclrfinalcopy
\title{From Reward Signal to Visual Utility:\\A Controlled Audit of\\Medical VLM Post-Training}
\author{Wang Jingxin\\Institute of Neuroscience, CAS\\\texttt{wangjx7777777@gmail.com}}
\begin{document}
\maketitle
\lhead{}
\begin{abstract}

Medical vision-language model (VLM) post-training is commonly evaluated through answer accuracy. We examine how changes in accuracy and training objectives relate to image-conditioned decisions in a controlled Qwen2.5-VL-3B study on PMC-VQA. We compare supervised fine-tuning (SFT) with low-rank adaptation (LoRA) restricted to the language model, expanded multimodal adaptation scopes, standard answer-only Group Relative Policy Optimization (GRPO), and a counterfactual evidence objective. On 2,000 clean-test questions, language-model LoRA SFT changes correct-image accuracy by +1.10 percentage points (95\% paired bootstrap CI: \mbox{-0.85} to +3.05), while visual-benefit events decrease by 2.40 points and image sensitivity decreases by 5.60 points. Paired records reveal 155 acquired and 203 lost visual-benefit events. Broader adaptation yields lower correct-image accuracy than language-model LoRA SFT. Standard GRPO produces mixed-reward groups and parameter updates, with an uncertain clean-test accuracy change. A generation audit reveals that canonical option scores can follow a different token path from generated answers. With scores taken along the greedy generation path, the evidence target improves on the training set; its gains over standard GRPO remain inconsistent on validation data at matched training doses. Sample-level analyses trace how evidence scores, decision margins, and generated answers change during post-training. This empirical and measurement audit identifies gaps between optimization activity, target acquisition, and useful held-out visual behavior.

\end{abstract}

\FloatBarrier

\section{Introduction}

A medical VLM can answer a question using relevant visual evidence, information in the question, or learned answer priors. Post-training may change the balance among these sources while producing only a small change in average accuracy. Understanding this change requires examining both the answers and their dependence on the image. We study this relationship by following the same questions across training stages and controlled image interventions.

Shortcut learning provides a broader context for this problem \citep{geirhos2020shortcut}. In medical VLMs, MedVLThinker reports strong performance from text-only reinforcement learning with verifiable rewards (RLVR) \citep{huang2025medvlthinker}, while Beyond Accuracy uses image interventions to quantify beneficial and harmful image effects \citep{zafar2026beyond}. Building on these findings, we ask: how do post-training updates change the visual benefit of individual answers, and when do improvements in a training objective translate into better image-conditioned decisions?

These questions motivate analyses at several levels. A decrease in the visual-benefit rate can arise when answers become correct under both the original and replacement images, or when previously correct answers become incorrect. A gain in an evidence score can reflect stronger support for the correct-image answer, lower scores under counterfactual inputs, or additional training exposure. Paired answer transitions, score decompositions, and matched-dose controls allow us to examine these possibilities separately.

We use Qwen2.5-VL-3B \citep{bai2025qwen} and PMC-VQA \citep{zhang2023pmcvqa} to compare SFT parameter scopes and answer-only GRPO under fixed evaluation protocols. A 2,000-question clean-test set measures their effects on task accuracy and visual benefit. Training and validation subsets support a detailed analysis of a counterfactual evidence objective, including its token-level measurement and dose-matched comparison with standard GRPO. A separate native-answer experiment on SLAKE examines how answer-token learning relates to generation and termination.

The analysis yields three main findings. First, SFT produces substantial gains and losses in visual-benefit events despite a small net change in task accuracy. Second, broader multimodal adaptation lowers clean-test accuracy, while standard GRPO leaves accuracy nearly unchanged despite active optimization. Third, the interpretation of an evidence objective depends on the token path used to measure it, the amount of training, and the score components that change. Together, these findings connect aggregate post-training outcomes to the decisions made on individual questions.

\FloatBarrier

\section{Related work}

\subsection{Accuracy and intervention-defined visual utility}

Beyond Accuracy evaluates visual reliance using correct and shuffled images: its VRS measures the accuracy contrast, while VBR and VHR separate beneficial and harmful image effects \citep{zafar2026beyond}. We extend this evaluation to paired questions across post-training stages and parameter scopes. Tracking each question reveals which visual-benefit events are acquired, retained, or lost, and how those transitions combine into aggregate changes.

CORAL combines hard-negative image substitutions with a contrastive grounding objective and reports gains for a 7B model \citep{zafar2026coral}. Its use of counterfactual images motivates examining how an image-dependent training signal affects subsequent decisions. Our study follows a counterfactual evidence objective on a 3B model through score measurement, training-target acquisition, and dose-matched validation, relating changes in the objective to changes in answer selection.

\subsection{Parameter-efficient adaptation and medical RL}

Zhou et al. report benefits from connector adaptation in multimodal parameter-efficient fine-tuning \citep{zhou2024peft}. This motivates our comparison of language-model LoRA, language LoRA with merger tuning, and an additional vision-LoRA configuration. Evaluating these configurations on the same questions links adaptation scope to task accuracy and changes in visual benefit.

MedVLThinker compares distilled-reasoning SFT and RLVR \citep{huang2025medvlthinker}, and Zhu et al. examine initialization, medical semantic alignment, and answer length in medical RL \citep{zhu2026medicalrl}. These studies highlight the importance of training data, initialization, and reward design. We examine answer-only GRPO with LoRA by measuring within-prompt reward variation, verifying parameter updates, and comparing the resulting checkpoint with its SFT initialization on held-out questions.

\subsection{Scoring faithfulness and grounding-oriented objectives}

Wang et al. show that first-token evaluation can disagree with instruction-tuned text outputs \citep{wang2024myanswer}, and Sanz-Guerrero et al. demonstrate the effect of answer-boundary tokenization on multiple-choice scoring \citep{sanz2025mind}. We examine the implications for optimization: when option scores define an evidence objective, the token path used to extract those scores affects both the objective and its interpretation. We therefore compare canonical option scoring with scores recorded along actual multimodal generation trajectories.

Grounding-oriented objectives supervise several aspects of visual reasoning. PAPO uses a perception-oriented KL objective \citep{wang2025papo}, Ground-R1 addresses region-scale reward imbalance \citep{cao2025ground}, and Act as you think develops perceptual and logical consistency rewards for medical VQA \citep{jiang2026act}. Our counterfactual objective uses the gold-answer score difference between correct and altered image conditions. A standard-GRPO control at the same training dose evaluates its incremental effect, while score decomposition relates that effect to the correct-image decision margin.

\FloatBarrier

\section{Experimental design and definitions}

\subsection{Study design, data, and model variants}

We organize the PMC-VQA experiments around three questions: how SFT changes image-conditioned answers, whether expanding the trainable components changes those effects, and whether a counterfactual evidence objective improves on answer-only GRPO at the same training dose. Qwen2.5-VL-3B-Instruct provides the common backbone. SFT uses 10,000 training questions and a one-epoch schedule; 1,500 validation questions are used for checkpoint selection. The clean-test set contains 2,000 questions associated with 1,440 distinct image hashes. Exact image-content and image-question comparisons found no overlap between the training and clean-test cohorts.

We denote the backbone before task-specific post-training as M0. In supervised fine-tuning, the model learns to generate dataset answers from image-question inputs. We compare three configurations that vary the trainable components. M1-L applies LoRA \citep{hu2021lora} to the language-model attention projections, keeping the vision encoder and multimodal merger frozen. M1-ML trains the merger together with the language adapters. M1-VML additionally applies LoRA to the vision encoder. This comparison tests whether adapting more of the visual pathway improves the same held-out endpoints. Checkpoints are selected by validation accuracy under the correct-image condition, with parse failures and the earlier checkpoint used to break statistical ties according to the fixed selection rule. Visual-benefit measures are evaluated separately from checkpoint selection. All three selected SFT checkpoints occur at half an epoch.

We then initialize answer-only GRPO from M1-L and denote this training trajectory as M2. GRPO samples multiple answers per prompt and updates the policy using their relative correctness and format rewards. Of the checkpoints along the 2,000-prompt trajectory, validation selected the checkpoint after 500 prompt exposures, denoted M2-500, for clean-test evaluation. Two evidence-objective variants examine how the scoring implementation affects learning: M3 uses canonical option-token scores, and M4 uses scores aligned with the actual generation path. We compare M4 with M2 after 500, 1,000, and 2,000 prompt exposures to separate the effect of the evidence objective from the amount of training. These fixed-dose comparisons use training and validation diagnostic cohorts.

The clean-test comparisons evaluate the SFT configurations and the validation-selected standard-GRPO checkpoint. Evidence-objective analyses examine score acquisition and transfer on the training and validation subsets. Because the validation set was also used during objective development, these analyses are exploratory. Paired predictions support the transition counts and sample-level analyses reported below.

\subsection{Three image interventions}

Each question is evaluated under three conditions: Correct (C), using its original image; NoImage (N), omitting the visual input; and Shuffled (S), using a replacement image. The question, answer choices, and dataset gold label remain fixed. Replacement images are assigned by a fixed question-level mapping, so questions sharing a source image may receive different replacements. Paired comparisons across models use the same mapping.

Let C, N, and S also denote binary answer correctness for one question, and let $a_C$ and $a_S$ denote parsed option labels. The primary definitions are:

\begin{align}
\operatorname{Acc}_C &= \mathbb{E}[C],\quad
\operatorname{Acc}_N = \mathbb{E}[N],\quad
\operatorname{Acc}_S = \mathbb{E}[S], \\
\Delta\mathrm{Vision} &= \operatorname{Acc}_C-\operatorname{Acc}_S,
\quad \Delta_N = \operatorname{Acc}_C-\operatorname{Acc}_N, \\
\mathrm{VBR} &= P(C=1,S=0),\quad \mathrm{VHR} = P(C=0,S=1), \\
\Delta\mathrm{Vision} &= \mathrm{VBR}-\mathrm{VHR},\quad
\mathrm{IS} = P(a_C\ne a_S), \\
\mathrm{VDS} &= P(C=1,N=0,S=0)=P(100), \\
\mathrm{LSR} &= P(C=1,N=1,S=1)=P(111).
\end{align}

VBR is the visual-benefit rate: the answer is correct with the original image and incorrect with the replacement. VHR is the visual-harm rate, with the reverse correctness pattern. Image sensitivity (IS) measures how often the parsed answer changes between these two conditions. The three-bit code records correctness in C/N/S order. VDS denotes strict vision-dependent success under the evaluated interventions (100), and LSR denotes success under all three conditions (111). Patterns 010 and 001 indicate success only under NoImage and Shuffled, respectively. These patterns summarize the responses of a particular model to the specified interventions.

\begin{figure}[!htbp]\centering
\includegraphics[width=\linewidth]{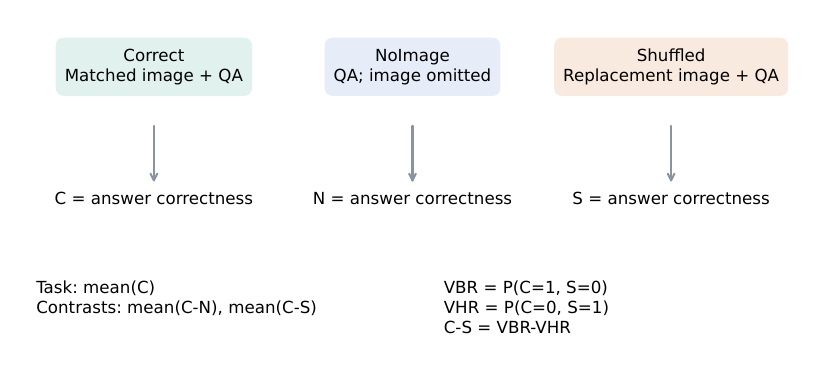}
\caption{Evaluation under Correct, NoImage, and Shuffled image conditions. The three-bit patterns record answer correctness in C/N/S order. Visual-benefit and visual-harm events decompose the Correct-minus-Shuffled accuracy difference; the full patterns also describe success under the NoImage condition.}\label{fig:1}
\end{figure}

\subsection{Training and inference controls}

For SFT with LoRA restricted to the language model, LoRA adapts 144 language-model q/k/v/o projection modules, totaling 7,372,800 trainable parameters. The rank is 16, alpha 32, and dropout 0.05. SFT uses a learning rate of 1e-4, effective batch size 8, warmup ratio 0.03, maximum gradient norm 1, BF16, and gradient checkpointing. The processor uses a fixed pixel range of 200,704 to 1,003,520. The three SFT configurations vary the trainable components as described above.

Standard GRPO uses within-prompt relative rewards \citep{shao2024deepseekmath}, with G = 4 completions per prompt, temperature 1.0, top-p 0.95, top-k 0, and a maximum of 32 new tokens. The learning rate is 1e-6 and beta is 0.04. The reward combines answer correctness with a 0.05-weighted format-compliance term. M1-L supplies the initialization and reference model. Training targets use the answer-only format \texttt{<answer>X</answer>}.

Validation and token-trace analyses use the tagged format \texttt{<answer>X</answer>}. Clean-test evaluation requests \texttt{Answer: X} followed by a brief reason, using greedy generation, a maximum of 128 new tokens, BF16, SDPA attention, and batch size one. Each endpoint applies its protocol consistently across the compared models. The generation-path analysis in Section 4.4 examines greedy decoding with the tagged-answer format.

The clean-test parser extracts the option label from the requested answer format and also accepts a completion consisting solely of A, B, C, or D. Outputs that fail parsing count as incorrect predictions. Applying this parser consistently gives an M0 correct-image accuracy of 51.70\%.

\subsection{Statistical analysis and case selection}

For the principal comparisons, model differences are summarized by paired question-level bootstrap intervals using 10,000 replicates and a fixed random seed. Resampling keeps each question's predictions paired across models and conditions. The intervals quantify sampling uncertainty for the evaluated checkpoints and cohort. We report effect sizes alongside the intervals; the implications of repeated images, training-seed variation, and multiple comparisons are discussed in the limitations.

We use paired predictions to count transitions between correctness patterns. Cases are selected post hoc as the lowest numerical sample index within each specified behavioral category, requiring valid parsed options for every displayed model and condition. Each case reports its selection category and pool size, together with dataset labels and model outputs. This rule makes the selection reproducible and allows the examples to be read alongside the aggregate transition counts.

\FloatBarrier

\section{Results}

\subsection{SFT redistributes success across questions and image conditions}

To examine how SFT changes performance on individual questions, we compare the base model (M0) with language-model LoRA adaptation (M1-L) on the same clean-test questions under all three image conditions. Table 1 reports the aggregate results. Correct-image accuracy rises from 51.70\% to 52.80\%, a paired gain of +1.10 percentage points (95\% CI [\mbox{-0.85}, +3.05]). The interval leaves the direction of the task effect uncertain.

\begin{table}[htbp]
\caption{Performance on the frozen PMC-VQA clean-test cohort of 2,000 QA. Values are percentages, except C-S, which is in percentage points. VDS and LSR correspond to patterns 100 and 111. Paired confidence intervals for model differences are reported in the text.}\label{tab:1}
\centering
\fontsize{8}{10}\selectfont\setlength{\tabcolsep}{2pt}
\begin{tabular*}{\linewidth}{@{\extracolsep{\fill}}lrrrrrrrr@{}}
\toprule
Model & Correct & NoImage & Shuffled & C-S & VBR & VHR & VDS/100 & LSR/111 \\
\midrule
M0 & 51.70 & 36.45 & 31.25 & 20.45 & 29.05 & 8.60 & 17.55 & 16.95 \\
M1-L & 52.80 & 36.20 & 34.55 & 18.25 & 26.65 & 8.40 & 17.60 & 20.15 \\
M1-ML & 51.25 & 36.40 & 34.15 & 17.10 & 25.35 & 8.25 & 16.20 & 20.45 \\
M1-VML & 51.15 & 36.85 & 33.85 & 17.30 & 25.00 & 7.70 & 16.50 & 20.70 \\
M2-500 & 52.60 & 36.70 & 34.50 & 18.10 & 26.60 & 8.50 & 17.35 & 20.25 \\
\bottomrule
\end{tabular*}
\end{table}

SFT also changes performance across image conditions. VBR falls by 2.40 points (CI [\mbox{-4.25}, \mbox{-0.55}]), IS by 5.60 points ([\mbox{-8.00}, \mbox{-3.20}]), and pattern111 rises by 3.20 points ([+1.40, +5.00]). Pattern100 changes by only +0.05 points ([\mbox{-1.65}, +1.75]). The two image contrasts move in opposite directions: C-S changes by \mbox{-2.20} points ([\mbox{-4.50}, +0.10]), while the C-N point estimate increases from 15.25 to 16.60 points. The apparent change in image utility therefore depends on the comparison condition.

\begin{figure}[!htbp]\centering
\includegraphics[width=\linewidth]{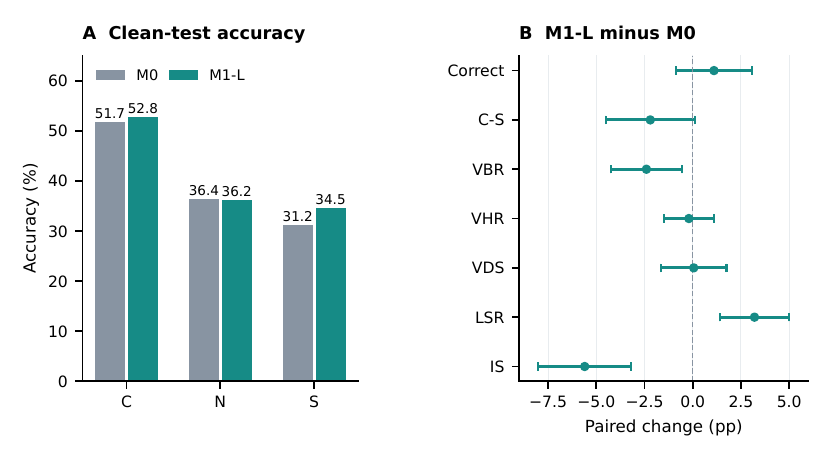}
\caption{SFT performance under the frozen clean-test protocol. Paired-bootstrap intervals show decreases in VBR and IS and an increase in pattern111, while the task and C-S changes remain uncertain. NoImage accuracy is nearly unchanged.}\label{fig:2}
\end{figure}

Paired transitions show the question-level changes underlying these averages. From M0 to M1-L, 155 visual-benefit events are gained and 203 are lost, yielding a net loss of 48/2,000. Correct-image success is gained on 209 questions and lost on 187, for a net increase of 22 correct answers. The nearly unchanged pattern100 count, from 351 to 352, likewise contains transitions into and out of the pattern.

\begin{figure}[!htbp]\centering
\includegraphics[width=\linewidth]{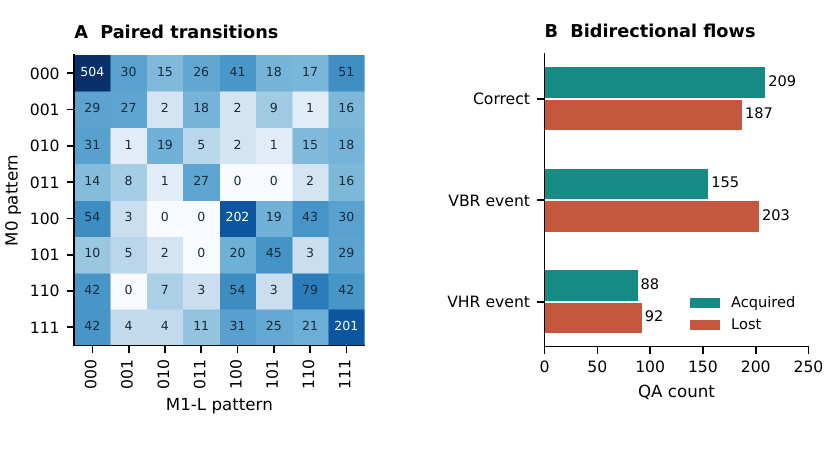}
\caption{Paired pattern transitions and bidirectional event flows from M0 to M1-L. Matrix entries are QA counts and sum to 2,000; color uses log(1 + count). The 155 VBR gains and 203 losses recover the aggregate \mbox{-2.40}-point change.}\label{fig:3}
\end{figure}

\textbf{Example transitions.} QA 000094 asks what panel A shows, with dataset gold Hydrocephalus. M0 predicts A/D/C under C/N/S, whereas M1-L predicts A/A/A, giving a 100-to-111 transition: correct-image success is retained and success extends to the other conditions. QA 000096 asks about panel C's yellow arrow on the same source image, with gold Chiari Malformation. M0 predicts D/B/A and M1-L predicts A/A/A, giving a 100-to-000 transition and a loss of correct-image success. These categories contain 29 and 54 parse-valid cases, respectively. The two questions use different panels and different shuffled replacement images.

The 38 parse-valid 000-to-100 cases include QA 000034, which asks whether a brown tumor is benign or malignant. M0 chooses the uncertainty option in all conditions; M1-L chooses the dataset gold, Benign, only with the correct image. This case illustrates newly acquired success specific to the correct-image condition, although the question wording also carries medical information. QA 000094 illustrates a separate output issue: M1-L's NoImage response says that image A shows hydrocephalus, making a visual assertion despite receiving no image.

\begin{figure}[!p]\centering
\includegraphics[width=\linewidth]{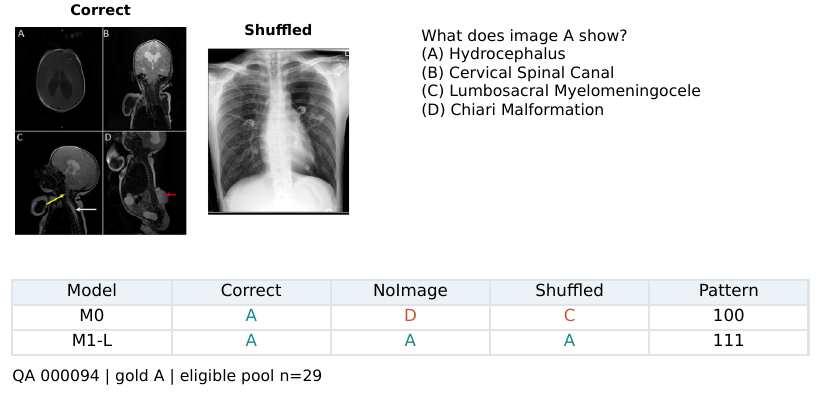}
\par\smallskip
\includegraphics[width=\linewidth]{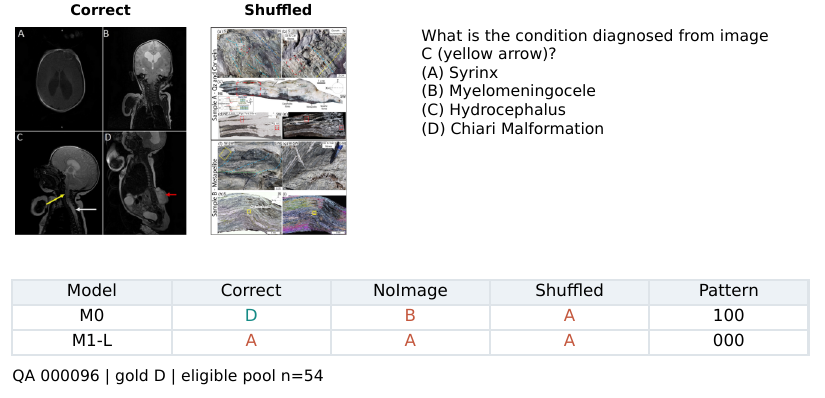}
\par\smallskip
\includegraphics[width=\linewidth]{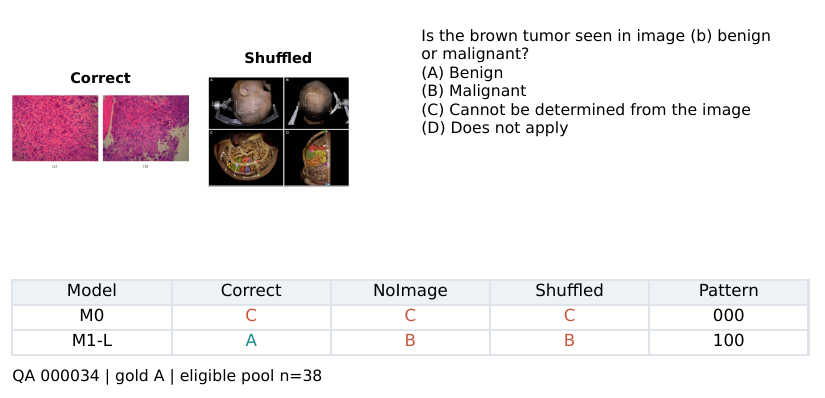}
\caption{Examples from three paired-transition categories. Labels follow the dataset annotations. The first two questions share a source image but concern different panels and use different replacement images. The third shows a gain in correct-image-only success.}\label{fig:4}
\end{figure}

Together, these transitions show how SFT redistributes success: some questions retain the correct-image answer while becoming answerable under other image conditions, some lose that answer, and others acquire it. Aggregate VBR and pattern111 changes summarize this redistribution, while the paired transitions reveal the different behaviors that produce it.

\FloatBarrier

\subsection{Expanding adaptation scope lowers accuracy in the tested SFT configurations}

To test whether updating the visual pathway changes the SFT outcome, we compare language-model LoRA (M1-L), language LoRA with merger tuning (M1-ML), and language and vision LoRA with merger tuning (M1-VML). This comparison follows evidence that connector tuning can improve multimodal PEFT performance \citep{zhou2024peft}. Relative to M1-L, Correct accuracy changes by \mbox{-1.55} points for M1-ML (CI [\mbox{-2.90}, \mbox{-0.25}]) and \mbox{-1.65} points for M1-VML ([\mbox{-3.05}, \mbox{-0.25}]). The difference between M1-VML and M1-ML is \mbox{-0.10} points ([\mbox{-1.50}, +1.30]). Both broader adaptation configurations therefore yield lower accuracy than language-model LoRA in this comparison.

The visual measures respond differently to the two scope expansions. M1-ML reduces VDS by 1.40 points relative to M1-L ([\mbox{-2.60}, \mbox{-0.25}]), while M1-VML reduces VBR by 1.65 points ([\mbox{-3.10}, \mbox{-0.20}]). Other differences have intervals crossing zero. The broader-scope models also have lower VHR point estimates, indicating fewer harmful image effects. Overall, the clean-test results reproduce only part of the VALID pattern.

\begin{figure}[H]\centering
\includegraphics[width=\linewidth]{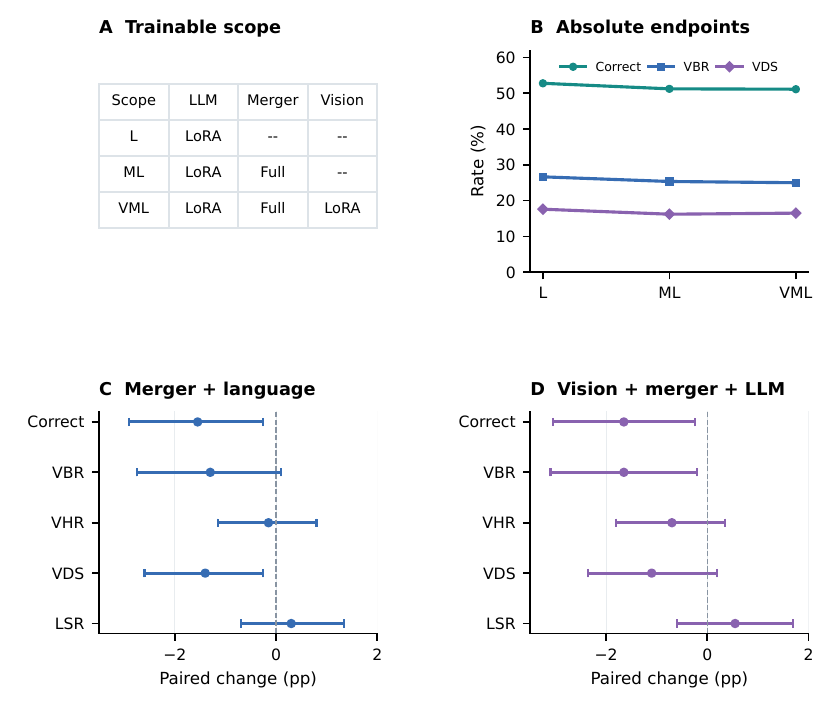}
\caption{Adaptation-scope comparison using VALID-selected checkpoints on a common clean-test cohort. The three configurations vary which components are trained: language attention adapters, the multimodal merger, and vision adapters. Their effects reflect the combined changes in trainable capacity, optimization, and regularization.}\label{fig:5}
\end{figure}

\textbf{Example transition.} QA 000070 asks which panel shows a full-thickness scaphoid fracture (gold: Image B). M1-L returns B/C/C, while both M1-ML and M1-VML return C/C/C, losing the correct-image answer. This is the lowest-index case among 62 parse-valid transitions from M1-L pattern100 to M1-VML pattern000 (Supplementary Figure S2). The outputs identify the answer change; its underlying cause could involve perception, panel binding, or answer selection.

\subsection{Standard GRPO produces reward variation and parameter updates with little change in accuracy}

To understand the small performance change after standard GRPO, we examine whether responses within each prompt group receive different rewards and whether training updates the adapters. Across the 0-500, 500-1,000, and 1,000-2,000 prompt segments, 71.2\%, 68.4\%, and 67.2\% of groups have nonzero reward variance. All 288 language-adapter tensors change during training. Thus, this run contains both within-group reward variation and parameter updates.

The selected M2-500 checkpoint reaches 52.60\% Correct accuracy on clean-test, compared with 52.80\% for M1-L, a paired change of \mbox{-0.20} points (CI [\mbox{-0.80}, +0.40]). VBR is 26.60\% versus 26.65\%, with 23 visual-benefit events gained and 24 lost. Correct-image success is gained on 17 questions and lost on 21. These opposing transitions account for the small aggregate changes.

\begin{figure}[!htbp]\centering
\includegraphics[width=\linewidth]{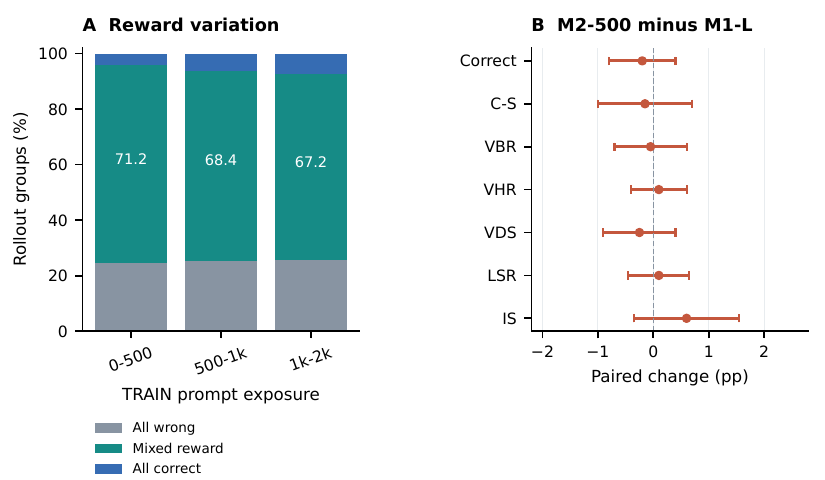}
\caption{Standard GRPO training diagnostics and clean-test performance. The training signal is the fraction of prompt groups with nonzero reward variance. A separate checkpoint audit verifies adapter updates. The clean-test comparison shows a small accuracy change with a paired confidence interval spanning zero.}\label{fig:6}
\end{figure}

\textbf{Example transitions.} QA 000016 changes from M1-L's A/B/B to M2-500's B/B/B, with gold B, Acute hepatitis. Correct-image success is gained, while the other two conditions remain correct. QA 000106 changes from C/C/C to D/C/C, with gold C, a scar on the dorsum of the left foot. Correct-image success is lost, while NoImage and Shuffled remain correct. These dataset-labeled examples are the lowest-index cases in the 17-gain and 21-loss pools.

A signal-stratified analysis also yields an uncertain high-versus-low signal learnability difference of \mbox{-0.0011} (CI [\mbox{-0.0106}, +0.0087]). Together, the reward, parameter, and learnability checks direct attention beyond signal scarcity to factors such as task difficulty and the relationship between the reward and held-out performance. The relative importance of these factors remains unresolved.

\subsection{Generation-path scoring changes the interpretation of the evidence objective}

We next examine whether the evidence score evaluates the answer branch used during generation. Tokenization and generation studies motivate checking this alignment \citep{wang2024myanswer,sanz2025mind}. Our canonical scorer used a branch with separate \texttt{>} and answer-letter tokens, whereas tagged generation used natural branch tokens such as \texttt{>A}. Across M1-L's 1,500 Correct-image VALID trajectories, the canonical score argmax disagrees with the generated answer on 906 questions (60.4\%). Disagreement rates are 76.4\% for NoImage and 66.7\% for Shuffled.

To reconstruct the generation branch, we retain token IDs, prefixes, multimodal state, processed scores, and termination tokens. The recorded traces cover 13,500 trajectories and 94,500 generated steps across M1-L, M2-500, and M3-500. The step-level check finds zero mismatches, and direct comparison with the raw traces finds zero parser-versus-frozen-answer mismatches. An independent A/B/C/D lexical tie-break selects a different answer on 473 trajectories (3.50\%); in every case, the actual answer remains in the highest-scoring tie set under the recorded tolerance. These checks establish step-trace consistency and account for ties in the branch scores.

\textbf{Generation trace.} VALID sample pmc\_train\_1 generates token IDs [27, 9217, 23465, 522, 9217, 29, 151645], corresponding to \texttt{<}, \texttt{answer}, \texttt{>A}, \texttt{</}, \texttt{answer}, \texttt{>}, and the terminal token. Re-tokenizing the displayed text omits the terminal token, so reconstructing the complete sequence and its decision branch requires the original generated tokens.

Scoring the actual generation branch changes the result for the M3 evidence objective. M3-500 minus M2-500 is \mbox{-0.037234} on VALID pattern100 and \mbox{-0.015625} on pattern110, measured in raw score units. The improved evidence alignment suggested by the canonical scorer therefore disappears under generation-aligned measurement.

\begin{figure}[H]\centering
\includegraphics[width=\linewidth]{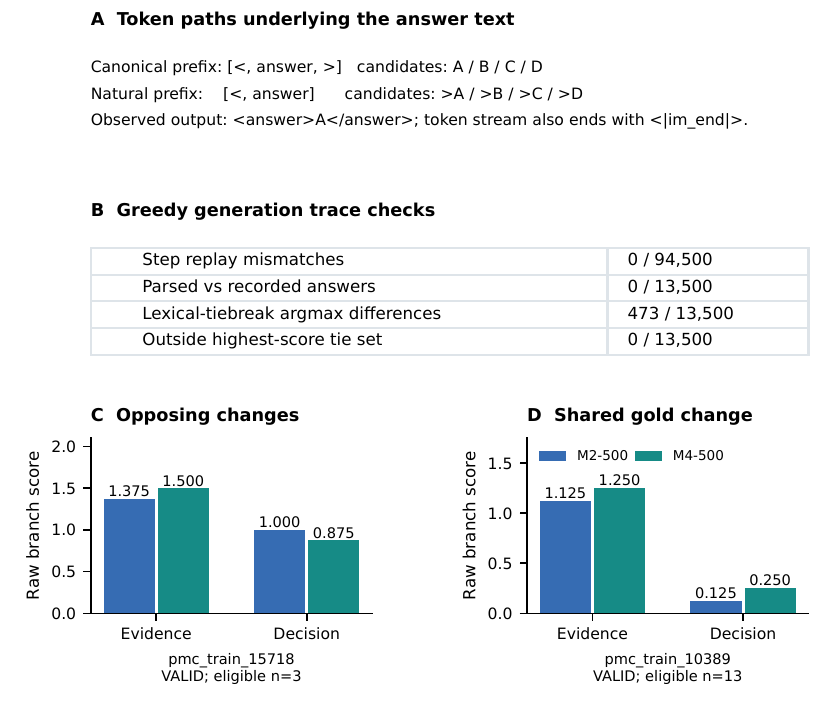}
\caption{Generation-path measurement and decision margins. Panels A-B show the natural token branch and trace-consistency checks. Panels C-D compare M4-500 with M2-500 on two VALID examples: evidence and decision margins can move in different directions or share a change in the correct-image gold score. The evidence-up/decision-down example is drawn from three cases meeting both strict criteria.}\label{fig:7}
\end{figure}

The generation-path audit changes the interpretation of evidence optimization because the canonical scorer evaluated a different answer branch. Aligning the score with generation provides the measurement basis for the following analysis of training-target acquisition, held-out transfer, and decision margins. The resulting cross-condition scores are expressed in raw branch-logit units.

\subsection{The evidence objective improves TRAIN scores, with uncertain matched-dose transfer}

Motivated by grounding-aware training \citep{zafar2026coral,wang2025papo,cao2025ground,jiang2026act}, we test whether an auxiliary evidence objective improves cross-condition score separation and whether those gains transfer at matched training doses. Let $\ell_C$, $\ell_N$, and $\ell_S$ denote the processed natural-branch gold scores. For frozen pattern100 questions, the target margin is $d=\ell_C-\max(\ell_N,\ell_S)$; for pattern110, it is $\ell_C-\ell_S$. M4 adds a 0.1-weighted softplus penalty on the negative margin, with pattern weights 1 and 0.5, respectively. Other patterns receive no evidence term. The objective operates on raw candidate-branch logits.

Relative to the initial M1-L checkpoint, mean TRAIN pattern100 target gains are +0.213867 at 500 prompts, +0.366699 at 1,000, and +0.428711 at 2,000 on the frozen diagnostic cohort (n = 256). These gains show that training increases the target score. To separate the effect of the objective from training exposure, we then compare M4 and standard GRPO at equal prompt counts.

On VALID pattern100 (n = 188), the matched-dose M4-k minus M2-k difference is \mbox{-0.017952} at k = 500 (CI [\mbox{-0.033910}, \mbox{-0.002660}]), \mbox{-0.000665} at 1,000 ([\mbox{-0.017952}, +0.015957]), and +0.013298 at 2,000 ([\mbox{-0.003989}, +0.031250]). Pattern110 effects (n = 112) remain negative in point estimate, with intervals crossing zero. On the diagnostic VALID cohort (n = 645), Correct, VBR, VHR, VDS, and C-S show no stable advantage for the evidence objective across doses. At 2,000 prompts, pattern111 decreases by 1.7054 points (CI [\mbox{-3.2558}, \mbox{-0.3101}]); this shift occurs alongside the uncertain task and visual-benefit effects.

\begin{figure}[!t]\centering
\includegraphics[width=\linewidth]{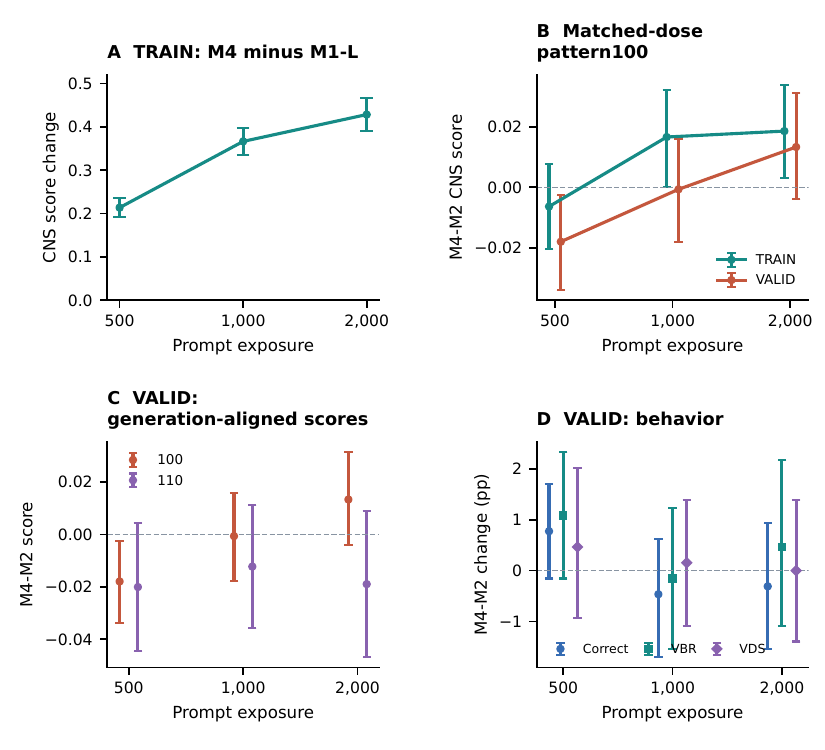}
\caption{Training-target acquisition and held-out transfer at matched doses. Panel A reports TRAIN score gains relative to M1-L. Panels B-D compare M4 with standard GRPO at equal prompt counts. The diagnostic cohorts contain 256 TRAIN pattern100 questions, 188 VALID pattern100 questions, and 112 VALID pattern110 questions; behavioral comparisons use the diagnostic VALID cohort of 645. Score effects are in raw branch-logit units and behavioral effects are in percentage points.}\label{fig:8}
\end{figure}

\Needspace{7\baselineskip}
To examine how evidence-score changes relate to answer selection, we compare the cross-condition evidence margin with the correct-image decision margin, $q=\ell_C(g)-\max_{w\ne g}\ell_C(w)$. Their changes are:

\begin{align}
\Delta d &= \Delta\ell_C(g)-\Delta b_{\rm cf}, \\
\Delta q &= \Delta\ell_C(g)-\Delta\max_{w\ne g}\ell_C(w).
\end{align}
Here $g$ denotes the gold option and $b_{\rm cf}$ the counterfactual gold-score baseline.

\textbf{Margin decomposition.} VALID sample pmc\_train\_15718 asks which technique produced micrographs C, D, and E (gold: TEM). From M2-500 to M4-500, d rises from 1.375 to 1.500 while q falls from 1.000 to 0.875. The correct-image gold score is unchanged: the counterfactual gold baseline falls by 0.125, and the strongest incorrect option under the correct image rises by 0.125. The answer remains unchanged. This case shows how the two margins can move in opposite directions through their different comparison terms.

The complementary sample pmc\_train\_10389 asks about pre-operative abdominal imaging (gold: Ultrasound and CT scan). Both d and q rise by 0.125 through an increase in the correct-image gold score, with the comparison terms unchanged. Across the 188-question pattern100 cohort, the Spearman association between evidence-margin and decision-margin changes is 0.2542 (CI [0.1321, 0.3683]). The exploratory residualized association after controlling for the correct-image gold-score change is \mbox{-0.0689} ([\mbox{-0.2130}, 0.0747]). This analysis examines the contribution of the gold-score term shared by the two margins.

The matched-dose comparisons show that the evidence objective's incremental validation benefit is inconsistent across training exposures. Read together with the score decomposition, the results locate two gaps between training and useful behavior: transfer of the learned target to validation questions, and translation of cross-condition score changes into a larger correct-image decision margin.

\subsection{Native-answer training improves gold-answer likelihood but disrupts output completion}

A separate SLAKE experiment \citep{liu2021slake} examines how answer acquisition under teacher forcing relates to free generation, a distinction studied in sequence learning \citep{ranzato2015sequence}. This experiment compares a native short-answer base model (R0) with its SFT variant (R1), using the likelihood of gold answers and the structure of generated completions to diagnose training behavior.

On 4,918 SLAKE TRAIN QA, R0-to-R1 normalized exact match falls from 56.57\% to 2.34\%, while mean gold-answer teacher-forced NLL/token improves from 4.3688 to 0.3759. VALID NLL also improves, from 4.3249 to 0.4831 on 1,053 QA. Gold-answer likelihood therefore improves on both splits even as free-generation accuracy falls on TRAIN. R1 emits comma-separated lists on 97.25\% of TRAIN questions, and 2.44\% of outputs contain a termination token. A post-hoc check finds that the first output segment exactly matches the gold answer in 71.88\% of cases, pointing to output completion as an important part of the failure.

\textbf{Completion example.} For slake\_train\_0 (gold: MRI), R0 outputs \texttt{MRI} and terminates. R1 outputs \texttt{mri, mri, mri, mri, mri, mri, mri, mri, mri, mri, mri} and reaches the 32-token cap without EOS. Its gold-answer NLL/token improves from 3.5234 to 0.0162, while termination NLL worsens from 0.0112 to 6.5600. Inspection confirms valid semantic labels, disabled packing, and matching templates across training and generation. The SFT targets omit termination supervision, identifying a potential contributor to the repeated completions. Its causal role remains unresolved.

The SLAKE results reveal a separation between learning the answer content and completing a usable response. The low gold-answer NLL, frequent correct first segments, and repeated continuations make termination behavior central to interpreting this branch. These observations motivate evaluating answer likelihood and termination alongside free-generation accuracy in native-answer post-training.

\FloatBarrier

\section{Discussion}

The experiments trace post-training effects from the learning signal to individual answers. Reward variation and parameter updates verify that optimization is active; generation-aligned traces connect the measured objective to the response path. Training-set target acquisition and dose-matched validation then assess learning and transfer, while correct-image margins and paired transitions reveal how these changes affect decisions. Following this sequence helps locate where an apparent improvement ceases to translate into useful behavior.

SFT substantially redistributes individual successes and dependence on the image condition. Aggregate task improvement remains uncertain, C-N and C-S move differently, and the overall prevalence of pattern100 changes little. Paired transitions explain how these nearly stable averages coexist with substantial gains and losses on individual questions. Following the same questions through post-training therefore extends intervention-based evaluation from a checkpoint-level comparison to an account of behavioral change.

The evidence-objective analysis identifies a sensitivity in cross-condition raw-logit targets. Adding a condition-specific constant to all option logits preserves within-condition probabilities and decisions while changing the cross-condition score difference. Generation-path alignment ensures that the score is measured at the relevant response branch, but the target remains sensitive to this arbitrary shift. The result motivates shift-invariant evidence objectives whose changes can be related directly to correct-image decisions.

The parameter-scope and GRPO controls help interpret the limited held-out improvement. Expanding trainable scope, obtaining variation in sampled rewards, and improving training margins each address a plausible bottleneck, yet held-out utility remains inconsistent across the tested configurations. The individual cases show several forms of behavioral change, including panel-answer loss and a correct-image gain that brings the answer into agreement with the already correct NoImage and Shuffled answers. Distinguishing perception, panel binding, option priors, and optimization effects within these cases requires additional measurements beyond the three-bit response pattern.

\FloatBarrier

\section{Limitations}

\textbf{Experimental scope and uncertainty.} The controlled study uses one 3B backbone, PMC-VQA, and a limited set of training recipes. Bootstrap intervals quantify question-level sampling uncertainty conditional on the trained checkpoints. Because the clean-test set contains repeated images, these intervals may understate uncertainty from image clustering; variation across training seeds is also unmeasured. Endpoint comparisons are unadjusted for multiplicity, and the illustrative cases were selected post hoc.

\textbf{Adaptive development.} Evidence objectives were developed and examined repeatedly on the same VALID set. The matched-dose comparison therefore provides a diagnostic assessment of held-out behavior, with positive subset effects requiring confirmation on an independent cohort. M4 held-out evaluation uses this development validation set.

\textbf{Image interventions.} The PMC shuffled mapping operates at the question level, and replacement images can cross modalities or contain nonclinical figure content. Removing the image can also introduce distribution shift. C-N and C-S should therefore be interpreted as responses to these specific input changes; their clinical meaning depends on the question and replacement image.

\textbf{Endpoint and label validity.} Multiple-choice performance can depend on information in the options, while native-answer exact match is sensitive to lexical variation. The two protocols measure different aspects of answer quality. Labels and images were evaluated as released, without independent clinician adjudication. Clinical validity, including the correctness of generated explanations, requires separate assessment.

\textbf{Cross-dataset coverage.} Evidence beyond PMC-VQA consists of the SLAKE native-answer pipeline diagnostic and the VQA-Med-2019 zero-shot baseline. A matched post-training comparison across medical datasets remains an open evaluation question.

\FloatBarrier

\section{Conclusion}

Across the tested Qwen2.5-VL-3B post-training configurations on PMC-VQA, substantial changes in individual answers coexist with uncertain aggregate task improvement. SFT creates and removes visual-benefit events, broader adaptation lowers correct-image accuracy and has mixed effects on visual metrics, and standard GRPO leaves selected-checkpoint accuracy nearly unchanged despite active optimization. After generation-path alignment, the counterfactual evidence objective improves its training target, while dose-matched validation shows inconsistent incremental benefit. Tracing these changes through decision margins and paired answers reveals where progress in the training objective translates into useful image-conditioned behavior and where it stops.

\FloatBarrier

\bibliography{references}

\bibliographystyle{iclr2027_conference}

\clearpage

\appendix

\setcounter{figure}{0}

\renewcommand{\thefigure}{S\arabic{figure}}

\setcounter{table}{0}

\renewcommand{\thetable}{A\arabic{table}}

\section{Supplementary analyses}

\subsection{Pattern composition and additional cases}

\begin{figure}[!htbp]\centering
\includegraphics[width=\linewidth]{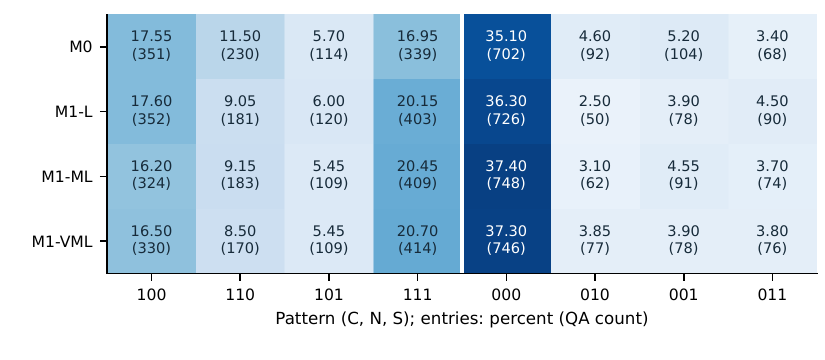}
\caption{Three-bit pattern distributions across the SFT parameter scopes. Rates are computed over 2,000 questions from 1,440 images. Figure 3 complements these aggregate distributions with paired question-level transitions.}\label{fig:S1}
\end{figure}

\begin{figure}[!htbp]\centering
\includegraphics[width=\linewidth]{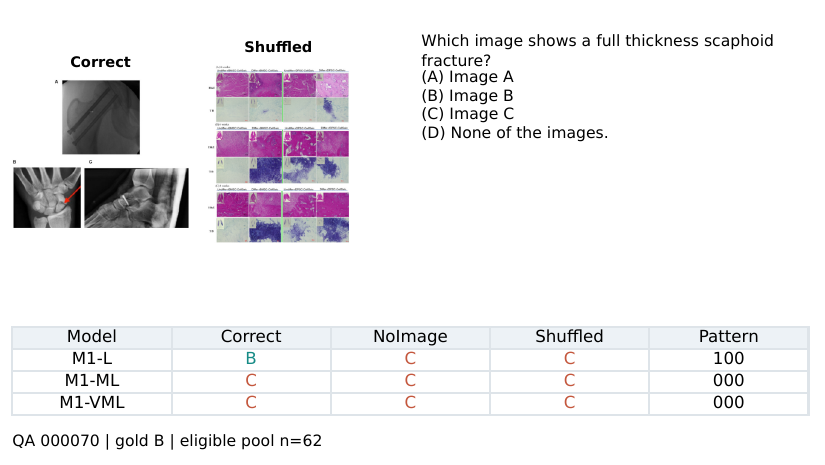}
\caption{Examples of answer changes under expanded adaptation and standard GRPO: loss of the correct panel answer, loss of the Correct-image foot-scar answer, and a Correct-image gain on an abdominal-CT question for which NoImage and Shuffled were already correct. Each case is the lowest-index member of its specified selection pool.}\label{fig:S2}
\end{figure}
\begin{figure}[!htbp]\ContinuedFloat\centering
\includegraphics[width=\linewidth]{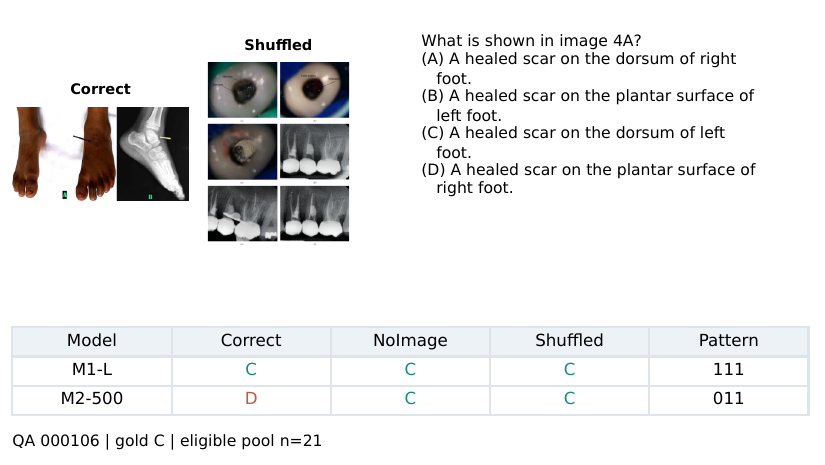}
\par\smallskip
\includegraphics[width=\linewidth]{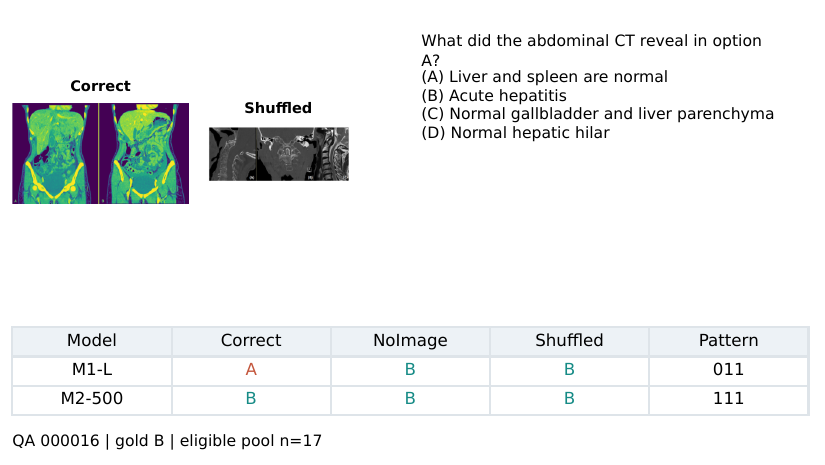}
\caption[]{(Continued.) In the abdominal-CT case, GRPO brings the Correct-image answer into agreement with the already correct NoImage and Shuffled answers.}
\end{figure}

\FloatBarrier

\subsection{Native-answer diagnostics and cross-dataset baselines}

\begin{figure}[!htbp]\centering
\includegraphics[width=\linewidth]{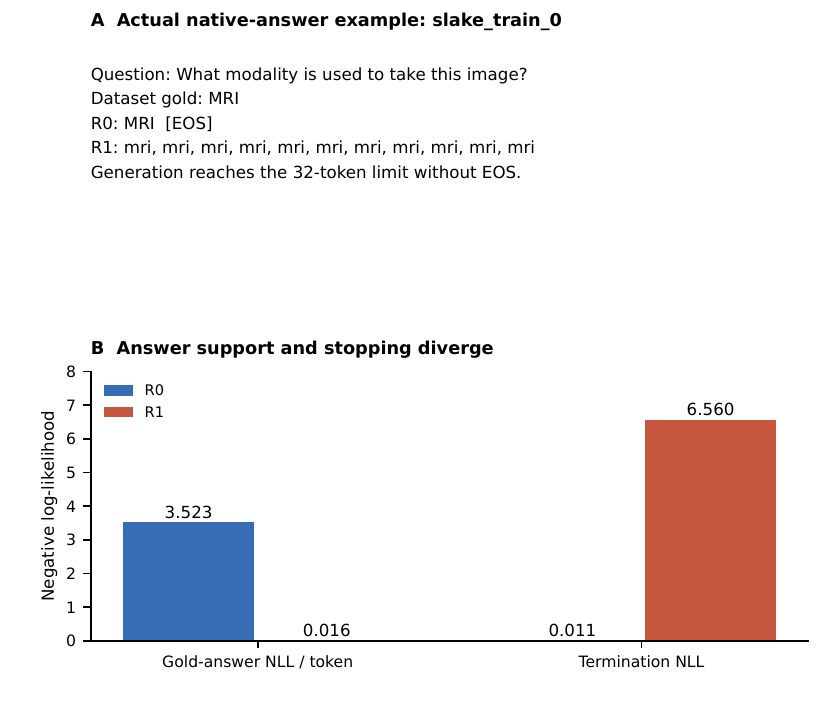}
\caption{Answer-token acquisition and stopping behavior in the SLAKE native-answer experiment. Separate scores for semantic answer tokens and EOS reveal improved answer-token likelihood alongside deteriorating termination behavior.}\label{fig:S3}
\end{figure}

\begin{table}[htbp]
\caption{Native-answer experiments on SLAKE and VQA-Med-2019 \citep{benabacha2019vqamed}. VQA-Med-2019 baseline C-N is +8.45 points (CI [7.00, 9.95]) and C-S is +3.55 points ([2.05, 5.05]). These results use each dataset's native-answer protocol. The SLAKE shuffled mapping contains 2 self-maps among 96 source-image groups, reducing the number of effective image replacements.}\label{tab:2}
\centering\small
\begin{tabularx}{\linewidth}{@{}>{\raggedright\arraybackslash}X>{\raggedright\arraybackslash}X>{\raggedright\arraybackslash}X>{\raggedright\arraybackslash}X@{}}
\toprule
Experiment & Evaluation cohort & Measurement & Result \\
\midrule
SLAKE base model (R0) & VALID: 1,053 QA & Native-answer exact match & C 55.56\%, N 30.58\%, S 32.67\% \\
SLAKE native-answer SFT (R1) & TRAIN generation; 5,971 teacher-forced scores per model & Answer-token likelihood and termination & Answer likelihood improves while repetitive generation and termination failure emerge \\
VQA-Med-2019 base model (R0) & VALID: 2,000 QA / 500 images & Native-answer exact match & C 13.90\%, N 5.45\%, S 10.35\% \\
\bottomrule
\end{tabularx}
\end{table}

\FloatBarrier

\subsection{Scoring and data provenance}

M0's primary clean-test accuracy is 51.70\% under the relaxed answer parser. In the generation-trace audit of M1-L, M2-500, and M3-500, direct comparison of raw parsed answers with the saved outputs yields 0 mismatches among 13,500 trajectories. The 473 differences from strict lexical argmax arise from tie handling. For the SLAKE native-answer experiment, normalized exact match is computed from the first nonempty output line. The comma-delimited first-segment score is reported separately as a diagnostic of answer acquisition during repetitive generation.

Case provenance is tracked through the frozen evaluation manifests. Selected images match their recorded SHA256 hashes, and split membership follows the manifest assignment. Some source identifiers retain the prefix \texttt{pmc\_train\_} after assignment to the repackaged VALID split.

\FloatBarrier

\Needspace{3.2in}

\subsection{Summary of experimental questions and findings}

\begin{table}[H]
\caption{Experimental questions, principal findings, and supporting analyses.}\label{tab:3}
\centering\small
\begin{tabularx}{\linewidth}{@{}>{\raggedright\arraybackslash}X>{\raggedright\arraybackslash}X>{\raggedright\arraybackslash}X@{}}
\toprule
Experimental question & Finding & Supporting analysis \\
\midrule
How much does the base model benefit from the correct image? & M0 C-S is +20.45 points & PMC clean-test image interventions \\
How does SFT change image-conditioned success? & Substantial question-level gains and losses coexist with uncertain aggregate task gain & Paired transitions and confidence intervals, Figures 2-4 \\
Does expanding trainable scope improve utility? & Broader adaptation lowers correct-image accuracy and has mixed effects on visual metrics & SFT scope comparison, Figure 5 \\
Do GRPO signals translate into selected-endpoint gains? & Selected clean-test accuracy changes by \mbox{-0.20} points despite active optimization & Signal audit and clean-test evaluation, Figure 6 \\
Does the canonical score follow the generated answer? & Generation-branch traces expose scoring-path discrepancies and tie-handling differences & Multimodal generation audit, Figure 7 \\
Does the aligned evidence objective acquire its training target? & The measured evidence target improves on TRAIN & Training-cohort diagnostic, Figure 8A \\
Does target acquisition improve dose-matched validation performance? & Incremental benefit is inconsistent across the tested doses & M4-k versus M2-k, Figure 8B-D \\
How do evidence-score changes relate to correct-image decisions? & Cross-condition score changes can be driven by components with little effect on the correct-image decision & Component analysis and sample trace, Figure 7C-D \\
How does native-answer SFT affect generation? & Answer-token likelihood improves while repetition and termination failure emerge & Paired generation and teacher-forced scores, Figure S3 \\
\bottomrule
\end{tabularx}
\end{table}

\FloatBarrier

\end{document}